\pdfoutput=1
\documentclass[sigconf]{aamas} 
\usepackage{balance} % for balancing columns on the final page
\usepackage{graphicx}

\setcopyright{none}
\acmConference[ALA '23]{Proc.\@ of the Adaptive and Learning Agents Workshop (ALA 2023)}
{May 29-30, 2023}{London, UK, \url{https://alaworkshop2023.github.io/}}{Cruz, Hayes, Wang, Yates (eds.)}
\copyrightyear{2023}
\acmYear{2023}
\acmDOI{}
\acmPrice{}
\acmISBN{}
\acmSubmissionID{???}

\title[Learning to Learn Group Alignment: A Self-Tuning Credo Framework with Multiagent Teams]{Learning to Learn Group Alignment: A Self-Tuning Credo Framework with Multiagent Teams}

\author{David Radke\footnotemark[1]}
\affiliation{
  \institution{University of Waterloo}
  \city{Waterloo}
  \country{Canada}}
\email{dtradke@uwaterloo.ca}

\author{Kyle Tilbury\footnotemark[1]}
\affiliation{
  \institution{University of Waterloo}
  \city{Waterloo}
  \country{Canada}}
\email{ktilbury@uwaterloo.ca}

\begin{abstract}
Mixed incentives among a population with multiagent teams has been shown to have advantages over a fully cooperative system; however, discovering the best mixture of incentives or team structure is a difficult and dynamic problem.
We propose a framework where individual learning agents self-regulate their configuration of incentives through various parts of their reward function.
This work extends previous work by giving agents the ability to dynamically update their group alignment during learning and by allowing teammates to have different group alignment.
Our model builds on ideas from hierarchical reinforcement learning and meta-learning to learn the configuration of a reward function that supports the development of a behavioral policy.
We provide preliminary results in a commonly studied multiagent environment and find that agents can achieve better global outcomes by self-tuning their respective group alignment parameters. 
\end{abstract}

\newcommand{\BibTeX}{\rm B\kern-.05em{\sc i\kern-.025em b}\kern-.08em\TeX}

\begin{document}

\newcommand*{\ourmean}[1]{\overline{#1}}

\newcommand{\oursubsub}[1]{\subsubsection{#1}}

% optional macro used to highlight things that need changing
\newcommand{\optional}[1]{\textcolor{YellowGreen}{#1}}
% \newcommand{\optional}[1]{}

% todo macro used to highlight things that need changing
\newcommand{\todo}[1]{\textcolor{red}{\textbf{#1}}}

\newcommand{\done}[1]{\textcolor{orange}{\st{#1}}}

\newcommand{\addressreview}[1]{\textcolor{black}{#1}} %added by Dave
% \newcommand{\addressreview}[1]{\textcolor{olive}{#1}}
%\newcommand{\timtext}[1]{}

% oldtext macros
% This is meant to draw lines through the text but it doesn't
% seem to work everywhere
% \newcommand{\oldtext}[1]{\textcolor{cyan}{\st{#1}}}
% Use this to mark text in light blue for potential removal
\newcommand{\oldtext}[1]{\textcolor{cyan}{#1}}
% Use this definition to hide the old text (useful to see how long the paper
% would be if we remove the oldtext
% $\newcommand{\oldtext}[1]{}

\newcommand{\heading}[1]{\vspace{3pt}\noindent\textbf{#1 }}

\newcommand{\mysubsection}[1]{\vspace{3pt}\noindent\textbf{#1 }}

\newcommand{\newtext}[1]{\textcolor{black}{#1}} %added by Dave
% \newcommand{\newtext}[1]{\textcolor{blue}{#1}}
%\newcommand{\newtext}[1]{#1}

% list item with some vertical space removed
\newcommand{\vitem}{\vspace{-5pt}\item}

% Trying different ways to get the guidelines to stand out more
% \newenvironment{guideline}{\begin{framed}\emph{\bf GUIDELINE:} \it }{\end{framed}}
% This did not work very well, way too large a frame
% \newenvironment{guideline}{\begin{mdframed}\emph{\bf GUIDELINE:} \it }{\end{mdframed}}
\newenvironment{guideline}{\vspace{0pt} \noindent \hrulefill \\ \emph{\bf \textcolor{blue}{GUIDELINE:}} \it }{\\ \vspace{-5pt} \hrule}

\newcommand{\squishbegin}{
 \begin{list}{$\bullet$}
  { \setlength{\itemsep}{0pt}
     \setlength{\parsep}{1pt}
     \setlength{\topsep}{1pt}
     \setlength{\partopsep}{0pt}
     \setlength{\leftmargin}{1.5em}
     \setlength{\labelwidth}{1em}
     \setlength{\labelsep}{0.5em} 
  } 
}

\newcommand{\squishtwobegin}{
 \begin{list}{$-$}
  { \setlength{\itemsep}{1pt}
     \setlength{\parsep}{1pt}
     \setlength{\topsep}{1pt}
     \setlength{\partopsep}{0pt}
     \setlength{\leftmargin}{1.5em}
     \setlength{\labelwidth}{1em}
     \setlength{\labelsep}{0.5em} 
  } 
}

\newcommand{\squishend}{
  \end{list}  
}

\newcommand{\experiment}{\vspace*{4pt}\noindent\textbf{Experiment Setup:\hspace{0.4em}}}
\newcommand{\experimentend}{}

\newcommand{\moveup}{\vspace{-8pt}}
\newcommand{\movecaptionup}{\vspace{-20pt}}
\newcommand{\movecaptionuptab}{\vspace{-17pt}}
\newcommand{\colfigwidth}{0.90\columnwidth}

% NOTE!!! Labels must come after captions.

% btable #1 - location
% etable #1 - label, #2 - caption
\newcommand{\btable}[1]{\begin{table}[#1] \begin{center} }
\newcommand{\etable}[2]{\end{center} \vspace{-5pt} \caption{#2} \label{#1} \vspace{-15pt}\end{table}}

\newcommand{\wbtable}[1]{\begin{table*}[#1] \begin{center} }
\newcommand{\wetable}[2]{\end{center} \caption{#2} \label{#1} \end{table*}}

% Define a figure by specifying the size in the x dimension
% xfigure: #1 - location #2 - xsize, #3 - filename, #4 - label, #5 - caption
\newcommand{\xfigure}[5]{\begin{figure}[#1] \begin{center} \leavevmode \epsfxsize=#2 \epsfbox{#3} \end{center} \vspace{-12pt} \caption{#5} \label{#4} \end{figure}}

\newcommand{\xfigurewide}[5]{\begin{figure*}[#1] \moveup \begin{center} \leavevmode \epsfxsize=#2 \epsfbox{#3} \end{center} \movecaptionup \caption{#5} \label{#4} \end{figure*}}

% Define a figure by specifying the size in the x dimension
% yfigure: #1 - location #2 - ysize, #3 - filename, #4 - label, #5 - caption
%\newcommand{\yfigure}[5]{\begin{figure}[#1] \begin{center} \leavevmode \epsfysize=#2 \epsfbox{#3} \end{center} \vspace{-20pt} \caption{#5} \label{#4} \end{figure}}
\newcommand{\yfigure}[5]{\begin{figure}[#1] \begin{center} \leavevmode \epsfysize=#2 \epsfbox{#3} \end{center} \caption{#5} \label{#4} \end{figure}}

% Define a figure by specifying the size in the x dimension
% xyfigure: #1 - location #2 - xsize, #3 - ysize #4 - filename,
% #5 - label, #6 - caption
%\newcommand{\xyfigure}[6]{\begin{figure}[#1] \begin{center} \leavevmode \epsfxsize=#2 \epsfysize=#3 \epsfbox{#4} \end{center} \vspace{-20pt} \caption{#6} \label{#5} \end{figure}}
\newcommand{\xyfigure}[6]{\begin{figure}[#1] \begin{center} \leavevmode \epsfxsize=#2 \epsfysize=#3 \epsfbox{#4} \end{center} \caption{#6} \label{#5} \end{figure}}

\newcommand{\bfigure}[1]{\begin{figure}[#1]}
\newcommand{\efigure}[2]{\vspace{-8pt} \caption{#2} \label{#1} \end{figure}}

\newcommand{\Kyle}[1]{\textcolor{green}{Kyle: #1}}

%%% The following commands remove the headers in your paper. For final 
%%% papers, these will be inserted during the pagination process.

\pagestyle{fancy}
\fancyhead{}

%%% The next command prints the information defined in the preamble.

\maketitle 
\def\thefootnote{*}\footnotetext{These authors contributed equally to this work.}
\renewcommand*{\thefootnote}{\arabic{footnote}}
\setcounter{footnote}{0}
\newcommand{\inteam}{\nu}
\newcommand{\outteam}{\overline{\nu}}

\newcommand{\selfw}{\psi}
\newcommand{\teamw}{\phi}
\newcommand{\sysw}{\omega}

\section{Introduction}

Cooperation and teamwork are central to the success of many human endeavours.
% The ability to work in teams can magnify a group's abilities beyond the capabilities of any individual.
% Observed in both animal and human behavior, the ability to work in teams can magnify a group's abilities beyond the capabilities of any individual.
Recently, there has been increasing support for the study of cooperation and teams being central to the development of artificial intelligence (AI) and multiagent systems (MAS)~\cite{DafoeNature2021,Dafoe2020OpenPI}.
% Recently, there has been an increasing interest in making the study of cooperation central to the development of artificial intelligence (AI) and multiagent systems (MAS)~\cite{DafoeNature2021,Dafoe2020OpenPI}.
% However, recent work has shown that more favorable global joint policies can be achieved when a population is defined into multiple sub-teams instead of being a fully cooperative group~\cite{Durugkar2020BalancingIP,Radke2022Exploring}.
Similarly to humans, intelligent agents cooperating and working in teams can enhance their capabilities beyond those of a single agent.
However, recent work has shown that agents defined to be fully cooperative can be sub-optimal; agents that are not fully aligned with their teammates can achieve more globally favorable results~\cite{Durugkar2020BalancingIP,Radke2022Exploring}.
% However, agents are not guaranteed to be fully aligned with their teammates and recent work has shown that populations that are not fully cooperative can achieve more globally favorable results~\cite{Durugkar2020BalancingIP,Radke2022Exploring}.

This paper extends a recently proposed model, \emph{credo}~\cite{radke2022importance}. Credo regulates how an individual learning agent optimizes for multiple objectives in the presence of teams.
% The noun credo is defined as ``the beliefs or aims which guide someone's actions"~\cite{OxfordDictionary}.
Specifically, credo represents how much the agent optimizes for the goals of different groups they belong to: their individual goals, the goals of any teams they belong to, and the goals of the entire system.
In previous experiments within multiagent reinforcement learning (MARL) environments, the credo model showed that the best global outcomes for a population of agents were achieved when agents in a larger group were somewhat selfish or when agents were mostly aligned with a smaller sub-team, robust to some amount of selfishness.
% In previous experiments, the credo model showed that the highest mean population reward was generated when agents in a larger group were somewhat selfish or when agents were mostly aligned with a smaller sub-team, robust to some amount of selfishness.
While credo was predetermined and fixed in these past experiments, the results motivate the key research question this paper aims to address: can giving agents the ability to dynamically tune their credo allow them to learn favorable group alignments automatically?
% However, the results motivate an interesting research question: can giving agents the ability to dynamically tune their credo allow them to discover favorable team structure and alignment?

% In this paper, we conceptualize and provide a road map towards developing algorithms to address this research question.
In this paper, we conceptualize and provide a preliminary approach that enables agents to self-tune their credo.
We provide theoretical foundations as to the motivation behind self-tuning credo in the context of different team structures and group alignment.
Further, we detail our framework that endows agents with the ability to tune their own credo.
The framework borrows implementation concepts from hierarchical reinforcement learning (HRL) and meta-learning.
Conceptually, each agent has a credo-tuning policy and a behavioral policy to maintain the decentralized nature of individual learning agents.
The values of an agent's credo ultimately shapes their reward function, and thus, the optimization landscape of their behavioral policy.
The dual-layer structure is similar to high and low-level policies in HRL, while the credo-tuning policy learning to shape the optimization landscape for the behavioral policy reflects that of a meta-learning problem.

We present preliminary results in a widely studied MARL environment, the Cleanup Gridworld Game~\cite{SSDOpenSource}, and outline future plans for evaluation. 
We show that, when starting from a known sub-optimal group alignment (i.e., sub-optimal credo), agents that tune their respective credos with our framework move to a better group alignment and learn a more globally favorable joint policy. 
While favorable team structures and group alignments have been explored in our preliminary testing environment, we describe our plans to test our framework in an environment where these are not known.
The goal of this work is to enable agents to optimize their behaviors towards the various groups they belong to in any environment -- enabling agents to learn better joint policies while eliminating the need for researchers and practitioners to engineer specific team structures and credo parameters.
With this paper we make the following contributions:

\begin{itemize}
    \item We provide theoretical motivation behind dynamically tuning credo (Section~\ref{subsec:tuning_motivation}).

    \item We conceptualize an agent framework to allow agents to self-regulate their own individual credo (Section~\ref{subsec:tuning_framework}).

    \item We present preliminary results demonstrating the efficacy of our framework (Section~\ref{sec:results}) and outline future work (Section~\ref{sec:future_work}).
    
\end{itemize}

% In this section we provide motivation for allowing agents to self-tune their credo parameters (Section~\ref{subsec:tuning_motivation}) and detail our framework that allows agents to do this (Section~\ref{subsec:tuning_framework}).
\section{Preliminaries}
\label{sec:background}

We model our base environment as a stochastic game $\mathcal{G}=\langle \mathcal{N}, S, \\ \{A\}_{i\in N}, \{R\}_{i\in N}, P, \gamma, \Sigma \rangle.$
$\mathcal{N}$ is our set of all agents that learn online from experience (with size $N \in \mathbb{N}$) and $S$ is the state space, observable by all agents, where $s_i$ is a single state observed by agent $i$.
$A = A_1\times \ldots \times A_N$ is the joint action space for all agents where $A_{i}$ is the action space of agent $i$.
$R = R_1 \times \ldots \times R_N$ is the joint reward space for all agents where $R_{i}$ is the reward function of agent $i$ defined as $R_i: S \times A\times S \mapsto \mathbb{R}$, a real-numbered reward for taking an action in an initial state and resulting in the next state.
$P:S\times A\mapsto \Delta(S)$ represents the transition function which maps a state and  joint action into a next state with some probability and $\gamma$ represents the discount factor so that $0 \leq \gamma < 1$.
$\Sigma$ represents the policy space of all agents, and the policy of agent $i$ is represented as $\pi_i:S \mapsto A_{i}$ which specifies an action that the agent should take in an observed state.\footnote{We can also allow for randomized policies.}

% This paper builds on the credo model from past work~\cite{radke2022importance}.
% Credo is a model to regulate how much an agent optimizes for different reward components it has access to.
% \Kyle{We talk about what credo is in the intro, related work, and background here. This paper definitely hinges upon the reader having an idea of what credo is, but this might be excessive? Not sure, seems necessary in each spot it's at.}
We use ``common interest'' to refer to when agents share their reward and a \emph{team} is a set of individual agents that can have some degree of common interest for team-level goals.
Given a population, multiple teams with different preferences and interests that are not in zero-sum competition may co-exist.
The collection of all teams is referred to as a team \emph{structure}.
% A team \emph{structure} $\mathcal{T}$ is a partition of the population into teams, and 
We denote the set of all teams as $\mathcal{T}$, the teams agent $i$ belongs to as $\mathcal{T}_i$, and a specific team as $T_{i} \in \mathcal{T}_i$.

The credo model presented in~\cite{radke2022importance} relaxes the assumption that teammates are fully aligned though common interest to allow settings where agents may only partially optimize for a team's goal.
For example, an agent may optimize their policy for the performance of one or multiple teams, while also being somewhat oriented towards it's own personal goals.
This is done by decomposing an agent's reward function to be a combination of their individual environmental reward $IR_{i} = R_i$, the rewards $i$ receives from each team they belong to $TR_{i}^{T_i} \forall T_i \in \mathcal{T}_i$, and the reward $i$ receives from the system of $|N|$ agents $SR_i$.
$TR_{i}^{T_i}$ and $SR_i$ can be implemented with any function to aggregate and distribute rewards.

Each agent has a credo vector of parameters where the sum of all parameters is 1, represented $\mathbf{cr}_i = \langle \selfw_{i}, \teamw_{i}^{T_1}, \dots, \teamw_{i}^{T_{|\mathcal{T}|}}, \sysw_{i} \rangle$, where $\selfw$ is the credo parameter for $i$'s individual reward $IR_i$, $\teamw_{i}^{T_i}$ is the credo parameter for the reward $TR_{i}^{T_i}$ from team $T_i \in \mathcal{T}_i$, and $\sysw_i$ is the credo parameter for the reward $i$ receives from the system $SR_i$.
The parameter notation is organized by increasing order of group size, so that  $\mathbf{cr}_i = \langle$self$, \dots,$ teams$, \dots,$ system$ \rangle$, where |self| < |teams| $\leq$ |system|.
An agent's credo-based reward $R_i^{\mathbf{cr}}$ is a weighted combination of that agent's credo parameters and reward from that group.
Expanded in Section~\ref{subsec:tuning_framework}, in this work we implement functions for $TR_{i}^{T_i}$ and $SR_i$ specially designed for the self-tuning scenario that maintain consistency with the original implementation in~\cite{radke2022importance}.

% \Kyle{This section needs some disentangling of talking about the original credo as it's our work here and now vs talking about it as being previous work. When talking about past work it should be like ``We extend this previous work. The model in the previous work did this...'' and not ``We extend our past work. We did this...'' (even though it's always the same authors extending the previous work haha). Parts of this section very much read like the latter. I will think about this}
\section{Related Work}

% Humans are relatively unique, when compared to other species, in that we have developed with an inherent bias towards teamwork.
Humans have developed with an inherent bias towards teamwork.
However, humans are only able to reliably maintain social relationships with a maximum number of individuals, causing them to form smaller groups~\cite{dunbar1993coevolution}.
% However, similar to other species, humans are only able to reliably maintain social relationships with a maximum number of individuals to form smaller teams~\cite{dunbar1993coevolution}.
% To overcome these limitations in military, sports, and work, humans have developed ``teams-of-teams'' to reduce the complexity of social relationships and to examine between-team behavior at a higher level~\cite{DeChurch2010PerspectivesTW}.
Analyzing between-team behaviors is often done in organizational psychology (OP), focusing on the concept of social identification or people's perceptions of their goals~\cite{Porck2019SocialII,Sundstrom1990WorkTA}.
Team members may need to balance tendencies for their own personal goals with the goals of their team or the entire system~\cite{Wijnmaalen2019IntergroupBI,Carter2019BestPF}.
Humans are continuously learning; thus, this balance of how humans optimize for goals is likely to be dynamic instead of static.

In AI, the concept of multiple non-conflicting teams within a larger system has been primarily explored for task completion~\cite{Grosz1996CollaborativePF,Tambe1997TowardsFT}, and more recently been used in social dilemma scenarios~\cite{Radke2022Exploring}.
Furthermore, agents optimizing their behavior while balancing between personal and group-level goals has been of growing interest to the AI community~\cite{Durugkar2020BalancingIP,McKee2020SocialDA}.
One such example of this is ad hoc teamwork which relies on the ability to assess the goals of an individual or group to best optimize a cooperative utility function~\cite{Stone2010AdHA,Macke2021ExpectedVO}.

A previously proposed model, credo, considers how agents optimize for various goals in the context of multiple non-conflicting teams within a larger system~\cite{radke2022importance}.
Credo defines how agents optimize for various groups they belong to, namely themselves, any teams they belong to, and the entire system.
While credo showed how groups with mixed motives or some degree of selfishness can significantly outperform fully cooperative populations, all agents' credo parameters were initialized the same and kept constant throughout experiments.
Other work has studied concepts of dynamic reward sharing and the emergence of coordination; however, that work relied on random perturbations of reward sharing parameters and did not consider the existence of defined team structures~\cite{gemp2020d3c}.

In this paper,
% motivated by dynamic behavior and continuous learning in humans, 
we propose an agent framework where agents are able to self-tune their individual credo parameters for groups they belong to.
Our model builds on hierarchical reinforcement learning (HRL) concepts to define multiple policies within a single individual learning agent.
Given a population of credo-tuning agents, we hope to develop continuously evolving policies that overcome sub-optimal team definitions, recover favorable joint policies, and preserve cooperation across multiple learning entities.

\section{Self-Tuning Credo}

% \todo{Assume we have a fully cooperative population (i.e., one team)... give step through as more learning agents are added to this environment...}

In this section we provide motivation for allowing agents to self-tune their credo parameters (Section~\ref{subsec:tuning_motivation}) and detail our framework that allows agents to do this (Section~\ref{subsec:tuning_framework}).

\subsection{Motivation}
\label{subsec:tuning_motivation}

% Previous work has shown the importance of team structures on the joint policies that individual agents learn~\cite{Radke2022Exploring}.
The main motivation behind allowing agents to self-tune their credo parameters is to recover a more favorable joint policy despite a sub-optimal group environment.
% overcome a sub-optimal team definition to perform better credit assignment.
For example, Figure~\ref{fig:cleanup_results-reward}, adapted from the original credo paper~\cite{radke2022importance}, shows a 33\% increase in mean population reward in the Cleanup gridworld game when a population of six agents were 80\% cooperative and 20\% selfish compared to fully cooperative (Scenario 1).
Both scenarios highlighted in Figure~\ref{fig:cleanup_results-reward} achieve the highest mean population reward because agents tend to learn better joint policies under certain combinations of team structure and credo definitions.
% better divide labor better and generate more reward than larger teams due to their ability to coordinate with fewer teammates.

\begin{figure}[t]
    \centering
    \includegraphics[width=0.8\linewidth]{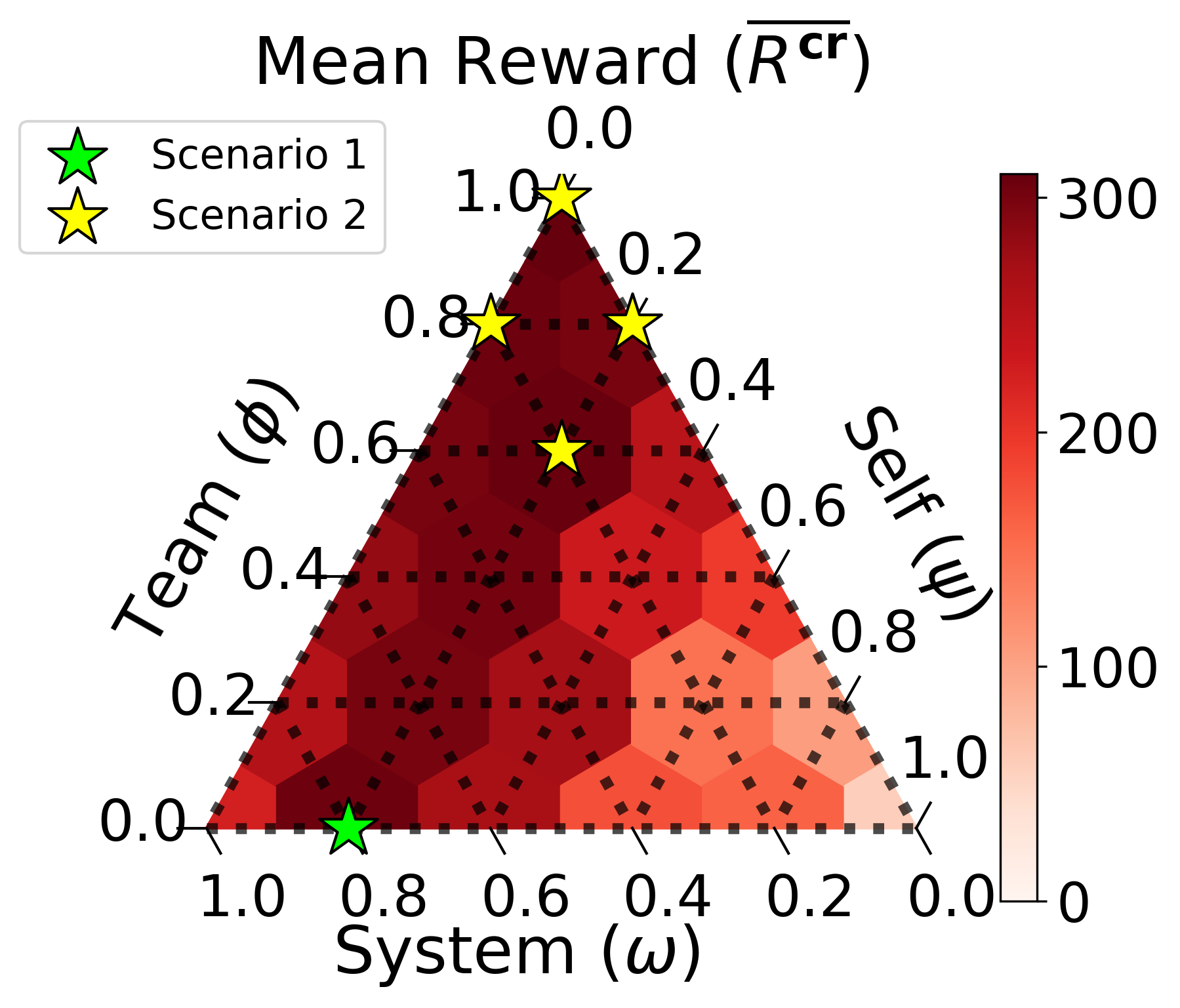}
    % \vspace{-20pt}
    \caption{Mean population reward for every credo parameter in the Cleanup environment from~\cite{radke2022importance}. These experiments have $|\mathcal{T}| = 3$ teams of two agents each. Two scenarios achieve the highest reward: when credo has slight self-focus paired with high system-focus (green star) and when team-focus is high (yellow stars).}
    \label{fig:cleanup_results-reward}
\end{figure}

Defining a team structure that best supports how individual agents learn is a difficult problem.
Recent work has used a fully cooperative population (i.e., shared reward function) to compare results with; however, the credo model has shown how a fully aligned population may be sub-optimal.
Providing agents with the ability to self-tune their credo parameters allows agents to regulate their internal reward function through group alignment.
For example, credo-tuning agents defined in one large cooperative population may discover the benefits of being slightly selfish on their own and converge to Scenario 1 in Figure~\ref{fig:cleanup_results-reward}.
The pressures to tune credo and recover different reward signals are highly correlated with the size of the reward-sharing group.
In this section, we detail how features of group size impact reward signals (Section~\ref{subsec:reward_group}) and how tuning credo can be leveraged to recover stronger reward signals (Section~\ref{subsec:recover_signal}).

% To understand \emph{why} it may be advantageous to allow agents agents to self-tune their credo vectors, 

\subsubsection{Reward Signals and Group Size}
\label{subsec:reward_group}

Consider a scenario with a fully aligned cooperative population of $N$ agents with only behavioral policies (i.e., only one group, the entire system).
% Assume an environment where non-zero rewards can be obtained by agents at any timestep of an episode; however, arbitrary movement 
In this setting, all agents share rewards at every timestep.
Thus, if agent $i$ collects a reward of $r$ at time $t$, all agents receive a reward of $\frac{r}{N}$ (assuming no other agent collects a non-zero reward from the environment).
The size of the reward-sharing group has two impacts on the reward function and agents' abilities to perform effective credit assignment.

% This has two impacts on agents' abilities to learn.
\textbf{Probability of non-zero reward approaches one:}
Starting with three common assumptions in reinforcement learning (RL), assume agents 1) are initialized with random policies, 2) fully explore the state space in the limit, and 3) each have equal probability of collecting a non-zero reward from the environment $P(r)_i$ (i.e., $P(r)_i = P(r)_j$ for any agents $i$ and $j$).
The probability of \textbf{any} agent collecting a non-zero reward is: $1 - P(r)^{N}$.
The derivative of this is positive, $f'(1 - P(r)^{N}) = N \cdot P(r)^{N-1}$; thus, agents in a reward-sharing group are monotonically more likely to receive a non-zero feedback signal at any timestep as the size of that group increases.
% since there exist more reward-sharing agents in the environment.
% Since agents are initialized to act randomly, the probability of any agent collecting a non-zero reward at any timestep increases monotonically with the size of $N$.
This probability approaches 1 as $N \rightarrow \infty$ in the limit.

\textbf{Variance of non-zero reward approaches zero:}
Agents receiving non-zero reward for their actions causes them to assign credit to these actions.
More positive reward for certain state-action pairs will result in them executing these state-action pairs more often in the future, and vice versa for negative reward.
However, while the \textbf{probability} of receiving a non-zero reward approaches 1 as $N$ increases, the derivative $f'\left( \frac{r}{N} \right) = - \frac{r}{N^2}$ implies the value of this non-zero reward monotonically approaches 0 as $N$ increases.
% Thus, the ability for agents to perform effective credit assignment for good or bad actions approaches zero as $N$ increases.
With a large group, the reward that each agent receives at every timestep will be a function of the expected number of agents that obtain rewards at any timestep.
% (i.e., $\frac{ar}{N}$).
As $N \rightarrow \infty$, the variance of this reward approaches zero.
Thus, agents would be unable to perform effective credit assignment if the size of their reward-sharing group is too large.

\subsubsection{Recovering a Stronger Reward Signal}
\label{subsec:recover_signal}

The previous subsection describes a scenario where agents lose the ability to perform effective credit assignment if the size of a reward sharing group is too large (assuming agents fully share rewards).
% That scenario assumes agents are fully aligned (i.e., fully system-focused).
% and are only able to learn from that single group-shared reward signal.
The credo model removes the assumption that agents fully share rewards to analyze situations where agents can learn from multiple types of groups they belong to.
Thus, regulating credo could allow agents to recover meaningful feedback signals from their actions in environments where credit assignment becomes challenging (i.e., if the reward-sharing group is too large).

% An example of credo-utilizing agents recovering a stronger feedback signal is present in the previously mentioned experiment of the original credo paper~\cite{radke2022importance}.
Consider again the results from the original credo paper shown in Figure~\ref{fig:cleanup_results-reward}.
Agents defined in a fully aligned population (one team of six agents) fail to converge to the most efficient joint policy; however, agents are able to recover a better joint policy when agents are 20\% selfish (Scenario 1).
% , their feedback reward signal becomes more responsive to their individual policies and agents converge to the best joint policy in the evaluation.
Agents that can self-regulate their credo parameters may recover better joint policies despite a sub-optimal environment that can impose credit assignment challenges, such as poorly defined team structures or group alignment.
% that would make credit assignment challenging without credo.
% While defining an optimal team structure is a difficult domain dependent problem, allowing agents to self-tune their internal credo vectors could help agents recover a stronger reward signal while still being exposed to the benefits of teams.

\subsection{Self-Tuning Credo Framework}
\label{subsec:tuning_framework}

% Meta-learning heirarchical type of thing.
% PPO three (or size of credo number of) continuous outputs for the credo agent.

This section details how we extend the credo-based reward function design and our proposed self-tuning credo framework.

\subsubsection{Extending Credo}

Recall from Section~\ref{sec:background} that agent $i$'s credo is defined as a vector of parameters that sum to 1, represented $\mathbf{cr}_i = \langle \selfw_{i}, \teamw_{i}^{T_1}, \dots, \teamw_{i}^{T_{|\mathcal{T}|}}, \sysw_{i} \rangle$, where $\selfw$ is the credo parameter for $i$'s individual reward $IR_i$, $\teamw_{i}^{T_i}$ is the credo parameter for the reward $TR_{i}^{T_i}$ from team $T_i \in \mathcal{T}_i$, and $\sysw_i$ is the credo parameter for the reward $i$ receives from the system $SR_i$.
In this paper, we define agent $i$'s credo-based reward function $R^{\mathbf{cr}}_i$ to be calculated as:

\begin{equation}
    R^{\mathbf{cr}}_i = \selfw_i IR_i + \sum_{T_i \in \mathcal{T}_{i}} \frac{\teamw_{i}^{T_i}}{\sum_{j\in T_i} \teamw_{j}^{T_i}} TR_{i}^{T_i} + \frac{\sysw_i}{\sum_{j\in N} \sysw_j} SR_i.
    \label{eq:credo_calc}
\end{equation}

Different from the original implementation, Equation~\ref{eq:credo_calc} allocates team and system rewards based on the \emph{ratio} of an agent's credo parameter for that group compared to the sum of credo parameters of other agents in that group.
This is necessary modification for the scenario when agents may have different credo parameters for the same group.
To maintain consistency with past work, we modify $TR_i^{T_i}$ and $SR_i$ to be the weighted sum of agents' rewards and their credo parameter for that specific group:
% \footnote{This is equivalent to the previous credo setting but expands to when teammates may not have the same credo for a team.}

\[ TR_i^{T_i}=\sum_{j\in T_i} \teamw_{j}^{T_i} R_j(S, A_j, S),\]

\[ SR_i=\sum_{j\in N} \sysw_{j} R_j(S, A_j, S).\]

This ensures all rewards that are collected from the environment are re-allocated to the various groups and scaled according to all credo parameters.
These modifications are equivalent to the previous credo setting when all agents have the same credo, but expand the reward function dynamics to when teammates may not have the same credo for a team.

\subsubsection{Agent Architecture}

An overview of our proposed agent framework is given in Figure~\ref{fig:framework_overview}.
The architecture of the agent is a multi-level policy inspired by HRL, where each layer influences the learning problem of the other.
% An agent operating within our proposed self-tuning credo framework is characterized by two policies. 
The ``low-level'' policy, $\pi_i$, is a typical behavioral policy that takes actions $a_i$ conditioned on an observed state $s_i$ within an environment.
At each timestep, rewards are shared with other agents according to the agent's credo parameters $\mathbf{cr}_i$.
% This is synonymous to a \emph{lower-level} policy in the HRL architecture.
% corresponding to agent $i$, which, as defined previously, specifies what action is to be taken in an observed state in the environment. 
The ``high-level'' policy, $\pi^{\mathbf{cr}}_i$, modifies the agent's credo parameters at a slower time scale.
Conditioned on the previous credo parameters, $\mathbf{cr}_i$, and the corresponding low-level policy's reward over $E \geq 1$ episodes, $\overline{R_i^{E}}$, the high-level agent produces updated credo parameters, $\mathbf{cr}_i^\prime$.
% for that agent in future episodes.
% The high-level agent having an action dimension of size $|\mathbf{cr}_i|$ draws inspiration from recent work on value function decomposition to infer the value of different reward parts~\cite{macglashan2022value}.
The top-layer policy operates at a slower time scale than the low-level behavioral policy to allow the low-level policy to gain experience with a particular credo and stabilize learning.
% before changing the credo parameters.
% This also prevents agents from dramatic credo parameter changes within an episode to stabilize learning and group assimilation.
% The second policy $\pi^{\mathbf{cr}}_i$, is the meta credo policy for agent $i$. The goal of this credo policy is to produce $cr{_i}^\prime$, an updated credo for agent $i$. 

% This is given the previous credo $cr{_i}$ and $R{_i}^{cr}$... where $R{_i}^{cr}$ is yielded by the behavioral policy from an action taken by the agent in the environment... This is for a single time step, but we using are the mean $R{_i}^{cr}$ across some number of time steps...

Both policies learn from experience using RL.
They both aim to individually maximize their sum of discounted future rewards and neither policy directly observes the other (i.e., they are both individual learning policies).
However, each policy directly influences the optimization landscape of the other.
The behavior of the low-level policy determines the reward feedback for the high-level policy; if the behavioral agent fails to gain reward, the high-level credo-tuning policy fails to get positive feedback.
Concurrently, the credo output of the high-level policy shapes the reward function of the low-level policy for the next set of $E$ episodes.

As mentioned in Section~\ref{subsec:reward_group}, tuning the amount of shared reward within groups regulates 1) the probability of an agent receiving a non-zero reward from a group with more teammates, and 2) the variance of their reward signal.
Thus, the high-level credo policy shapes the influence of these two aspects with respect to all groups referenced in the credo vector to guide the learning process of the low-level behavioral policy (self, any teams, and system).

\begin{figure}[t] %[!ht]
\centering
{\includegraphics[width=\columnwidth]{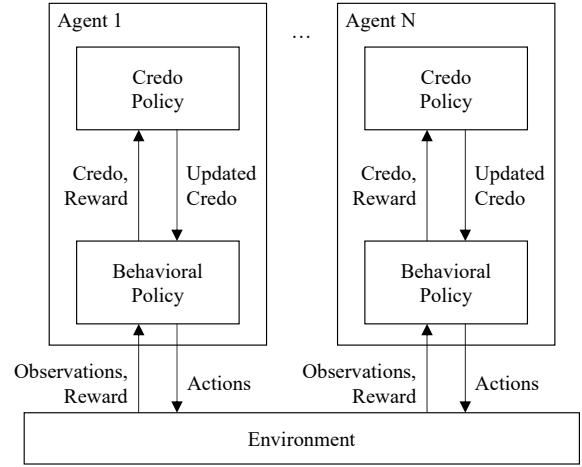} }%
\caption{Overview of the proposed credo-tuning agent framework. Each agent has two policies that operate at different time scales: a low-level behavioral policy that acts within an environment and a high-level credo-tuning policy that operates every $E \geq 1$ episodes. The credo-tuning policy shapes the optimization landscape for the behavioral policy while the learned behavior impacts the reward function for the credo-tuning policy.}%
\label{fig:framework_overview}%
\end{figure}

\section{Evaluation}

This section outlines our implementation details, experimental environment and setup, and presents preliminary experimental results. 

\subsection{Implementation}

\textbf{Low-level Behavioral Policy:}
We implement the behavioral policies of our agents with Proximal Policy Optimization (PPO)~\cite{PPO2017}. The PPO implementation in~\cite{radke2022importance} used an older version of the RLlib library (version 0.8.5) which made interconnecting the credo-tuning framework infeasible.
Thus, we adapted the same architecture as the agents in~\cite{radke2022importance} to the current version of RLlib (version 2.1.0) to incorporate the credo-tuning agent architecture shown in Figure~\ref{fig:framework_overview}.\footnote{\url{https://docs.ray.io/en/latest/rllib/index.html}}
% The updated version of the PPO algorithm performs slightly differently than the previous PPO implementation; thus, the absolute value of the reward obtained is not directly comparable to the results in~\cite{radke2022importance}.

\textbf{High-level Credo Policy:}
As a preliminary construction, and to reduce sample complexity, we implement the high-level credo policy as a $Q$-Learning agent with $\epsilon$-greedy exploration ($\epsilon = 20\%$)~\cite{watkins1992q}.
Consistent with the original credo paper, we define agents to belong to only one team, making credo vectors with three parameters (i.e., $\mathbf{cr}_i = \langle \selfw_i, \teamw_i, \sysw_i \rangle$).
We limit possible agent credos to intervals of 0.2, creating a state space of 21 possible states (shown in Figure~\ref{fig:cleanup_results-reward} from~\cite{radke2022importance}).
With three credo parameters, the agent can choose from any of seven actions.
The action space consists of either increasing/decreasing any combination of credo parameters (six actions) or doing nothing (one action).
For example, if $\mathbf{cr}_i = \langle 0.2, 0.0, 0.8 \rangle$, the agent can take an action to decrease self-focus and increase system focus (by increments of 0.2) to result in $\mathbf{cr}_i' = \langle 0.0, 0.0, 1.0 \rangle$.
If the agent chooses an action that would increase any credo parameter above 1.0 or below 0.0, no action is taken and $\mathbf{cr}_i = \mathbf{cr}_i'$.
The behavioral policies are updated with $\mathbf{cr}_i'$ for the next $E$ episodes.

\subsection{Environment}

We perform our preliminary evaluation in the Cleanup Gridworld Game~\cite{SSDOpenSource}.
% with behavioral policies adapted to RLlib (version 2.1.0).
% In future work, we plan to expand our analysis to include more trials to understand behavioral variance and include the Neural MMO environment (NMMO)~\cite{suarez2019neural}.
% \subsection{Cleanup Gridworld Game}
% \paragraph{Cleanup Gridworld Game:}
Cleanup is a temporally and spatially extended Markov game representing a sequential social dilemma.
% similar to past work~\cite{Jaques2019SocialIA,Hughes2018InequityAI}.
We keep the underlying environment unchanged from previous setups~\cite{Leibo2017MultiagentRL} except for the team and system reward functions.
Agent observability is limited to an egocentric 15 $\times$ 15 pixel window and consuming an apple yields +1 reward. 
Apple regrowth rate is dependent on the cleanliness of an adjacent river. 
To be successful in cleanup, agents must learn to balance actions of consuming apples and cleaning the river (which returns no positive reward).
Agent rewards are determined by their credo $\mathbf{cr}_i$ which is updated at regular intervals.
% 96 episodes (where there are six rollout workers with 16 environments per worker).
% We do change the reward function so that teammates share rewards, consistent with previous team-based work in Cleanup~\cite{Radke2022Exploring}.
% ; however, apple growth is conditional on the cleanliness of an adjacent river (cleaning yields no reward).
% Successful groups in Cleanup balance the temptation to free-ride and pick apples with the public obligation to clean the river.
% Prior research in this environment has shown that smaller teams divide labor better and achieve more reward than a larger fully cooperative system, but understanding why agents learn these behaviors has remained an open question~\cite{Radke2022Exploring}.
% Our work provides context to why this result emerges and evaluates team structures that learn efficient divisions of labor.
% Agents initialized in Cleanup come with a punishing action which returns negative rewards both when punishing or being punished.
% In our experiments, agents quickly learn to ignore using this action; therefore, we we omit these negative rewards from our theoretical analysis to prevent skewing our insights.
% We fix $N=6$ and modify the number $|\mathcal{T}|$ and size of teams $n$ since the complexity of the environment depends on the size of the population and to be consistent with previous work~\cite{Radke2022Exploring,radke2022importance}.
% A particular case of team structure is where agents belong to exactly one team.
Consistent with the original credo paper, we set the size of each team to be two agents $|T_i| = 2$, creating $|\mathcal{T}|=3$ disjoint teams from the population of $N=6$ agents~\cite{radke2022importance}.
% Formally, $\mathcal{T}$ is a partition of the population into disjoint teams,  $\mathcal{T} = \{ T_i | T_i \subseteq N, \cup T = N, T_i \cap T_j = \emptyset \forall i, j \}$.
Agents are implemented with PPO behavioral policies, $Q$-learning credo policies, and experiments last for $3.2 \times 10^8$ environment steps and credo parameters are updated every 96,000 environment steps.

\subsection{Experiment}
\label{subsec:experiment}

% \todo{Talk about things we want to investigate and experiement with...}
% We plan to investigate various credo initializations. Fully cooperative, moves towards optimal credo? Fully selfish? etc
% Different team configurations...

% \paragraph{Preliminary results.} Potentially show plot of how credos change during our `dumb' cleanup training run.

We design an experiment to evaluate if credo-tuning agents can overcome a sub-optimal initialization to recover a joint policy that achieves higher mean population rewards (Scenario 1 or 2 in Figure~\ref{fig:cleanup_results-reward}).
We initialize agents to be fully system-focused (i.e., $\mathbf{cr}_i = \langle 0.0, 0.0, 1.0 \rangle$).
The low-level behavioral policy trains, and the high-level credo policy updates $\mathbf{cr}_i$, every 96 episodes (rollouts of six workers with 16 environment copies each).
This is equivalent to initializing agents with credo parameters in the bottom left corner of Figure~\ref{fig:cleanup_results-reward}; however, agents' credo policies are now able to adjust the agent's credo parameters.
% The high-level credo policy learns to define parameters that shape the learning problem for the low-level behavioral policy, and the low-level behavior shapes the high-level credo policy problem space.

This setting directly evaluates our discussion in Section~\ref{subsec:reward_group}.
One of the key findings in previous work is that some amount of mixed incentives can achieve more favorable global outcomes than a fully cooperative population~\cite{radke2022importance,Durugkar2020BalancingIP}.
In Cleanup, agents that are slightly self-focused or fully team-focused (Scenario 1 and 2 respectively in Figure~\ref{fig:cleanup_results-reward}) learn a better global joint policy through division of labor.
In these settings, agents divide labor and learn to specialize into roles of four apple picking agents and two river cleaning agents.
When agents are fully system-focused, they specialize into the sub-optimal joint policy of three apple pickers and three river cleaners.

We hypothesize that a full system-focused group learns this sub-optimal joint policy due to more a difficult credit assignment problem.
Agents in Scenario 1 of Figure~\ref{fig:cleanup_results-reward} learn the best joint policy by recovering slightly stronger reward signals by being 20\% self-focused.
The design of this experiment evaluates the ability for credo-tuning agents to recover stronger reward signals and converge to a better joint policy.
Intuitively, this can be thought of as agents learning to configure their credo parameters such that they converge to high-reward areas of Figure~\ref{fig:cleanup_results-reward}.
\subsection{Preliminary Results}
\label{sec:results}

This section shows preliminary results of the experiment detailed in Section~\ref{subsec:experiment}.
In the credo tuning experiment, all agents are initialized in the Cleanup environment with $\mathbf{cr}_i = \langle 0.0, 0.0, 1.0 \rangle$.
Each agent's behavioral policy updates every 96,000 environmental timesteps (96 episodes), at which point the high-level credo policy modifies the agent's credo parameters.
The behavioral policy never observes the credo parameters but instead experiences changes to their reward function over the next batch of episodes.
We compare credo tuning to two configurations where credo remains static. In the static team-focus experiment, agents maintain $\mathbf{cr}_i = \langle 0.0, 1.0, 0.0 \rangle$ for the entire experiment and fully share rewards with their teammates.
In the static system-focus experiment, agents maintain $\mathbf{cr}_i = \langle 0.0, 0.0, 1.0 \rangle$ for the entire experiment and share rewards with all agents (i.e., a fully cooperative system).
In all experiments, there are six agents that are divided into three disjoint teams of two agents each (i.e., $N = 6$, $|\mathcal{T}| = 3$, and $|T_i| = 2$).
We execute four trials of each experiment configuration.

We observe the same patterns with the static experiments as in previous work~\cite{Radke2022Exploring,radke2022importance}: full team-focus performs significantly better than full system-focus.
However, we found that updating the PPO agents from RLlib 0.8.5 to RLlib 2.1.0 modified their learning curves so that agents learn more gradually (despite no changes to the algorithm configurations).
Thus, while our direct learning curves are not comparable to past work, the overall result remains consistent and we extend the duration of the experiments from $1.6 \times 10^8$ to $3.2 \times 10^8$ environment steps.

\subsubsection{Reward}

\begin{figure}[t]
    \centering
    \includegraphics[width=\linewidth]{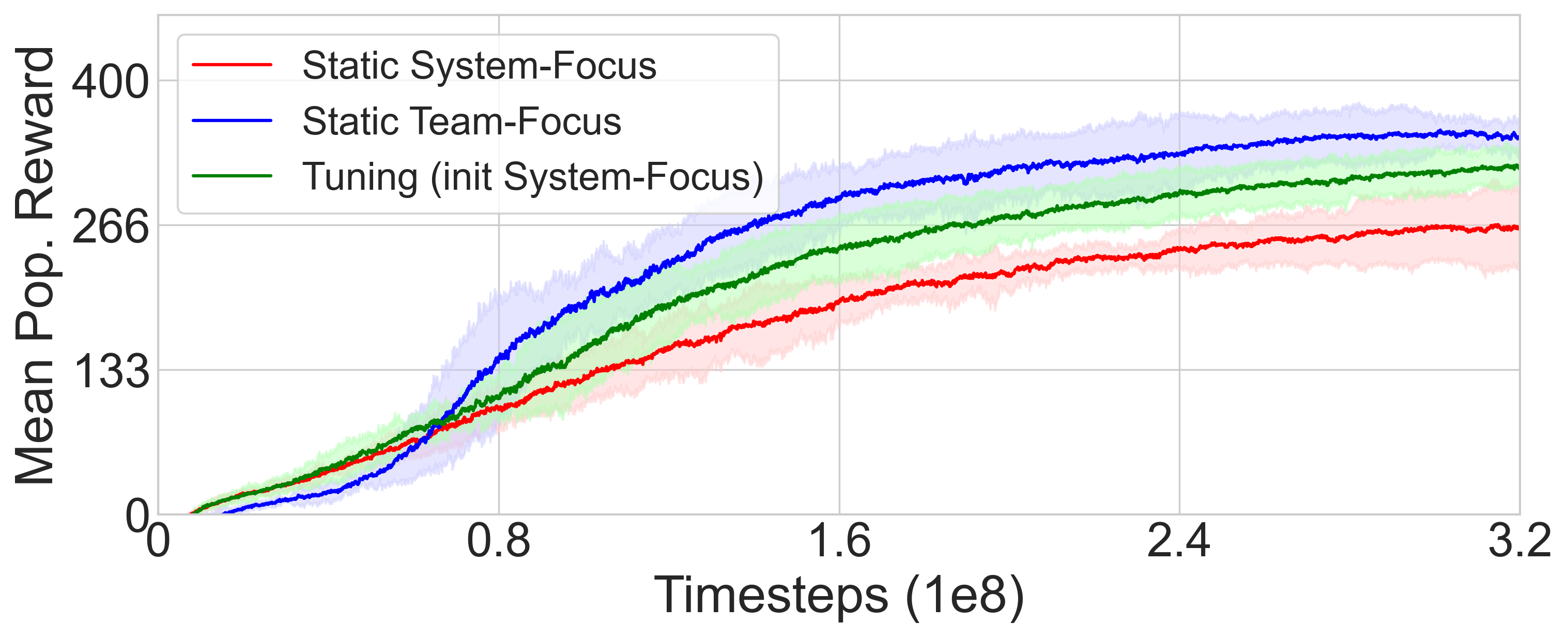}
    % \vspace{-20pt}
    \caption{Reward curves in the Cleanup environment for each experiment in our evaluation. Results are the mean across 4 trials for each experiment reported with 95\% confidence intervals. The static team-focus environment has been shown to achieve the highest mean population reward in Cleanup  with different credos (Figure~\ref{fig:cleanup_results-reward} Scenario 2). This shows that credo-tuning agents that are initialized with system-focus credo can increase their mean population reward to improve towards the level of team-focused agents.}
    \label{fig:reward_curve}
\end{figure}

\begin{figure}[t]
    \centering
    \includegraphics[width=\linewidth]{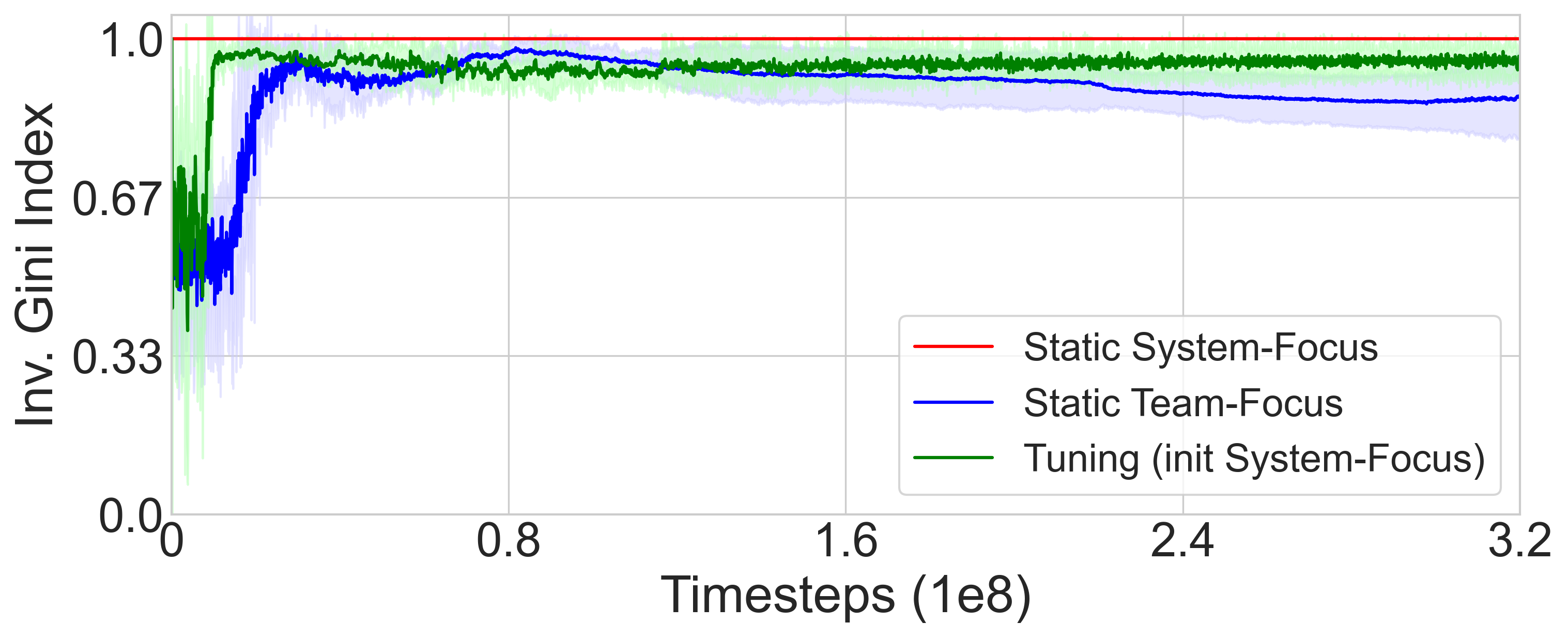}
    % \vspace{-20pt}
    \caption{Inverse Gini index curve in the Cleanup environment for each experiment in our evaluation. Results are the mean across 4 trials for each experiment reported with 95\% confidence intervals. Static system-focus credo is defined to have full equality and is always 1. This shows that credo-tuning agents converge to slightly higher equality than the static team-focused experiment.}
    \label{fig:gini_curve}
\end{figure}

Figure~\ref{fig:reward_curve} shows the mean population reward and 95\% confidence intervals obtained by the population of agents in the three different credo scenarios: static system-focus, static team-focus, and credo-tuning agents that were initialized to be system-focused.
The $y$-axis shows mean population reward and the $x$-axis shows timesteps of the experiment.
Consistent with past work, we find that static agents that are fully team-focused (blue) perform significantly better than static system-focused agents (red).
This is due to team-focused agents converging to a more efficient division of labor joint policy with two river cleaning agents and four apple picking agents, whereas system-focused agents converge to three agents each cleaning the river or picking apples.

Recall from Figure~\ref{fig:cleanup_results-reward} that full team-focus credo is one setting that achieves the highest reward in this configuration; thus, we treat full team-focus as an upper-bound result in this domain.
The goal of credo-tuning agents is not to overtake the team-focus credo, but converge to credo parameters that achieve higher reward than their initialized settings (i.e., fully system-focused credo; red line).
The green line in Figure~\ref{fig:reward_curve} shows the mean population reward and 95\% confidence intervals for the credo-tuning agents initialized with full system-focus credos.
Through the first 800,000 timesteps of the experiment, the credo-tuning agents (green) learn along the same trajectory as the system-focused agents (red).
However, giving agents the ability to modify their credo parameters leads to the population achieving roughly 21\% more mean population reward than the system-focus credo by the end of the experiment (320 for credo-tuning agents compared to 264 for static system-focus agents).
% \Kyle{Update this number with finalized results}
This shows the ability for credo-tuning agents to achieve more mean population reward despite a known sub-optimal team and credo parameter initialization.

\subsubsection{Reward Equality}

\begin{figure}[t]
    \centering
    \includegraphics[width=\linewidth]{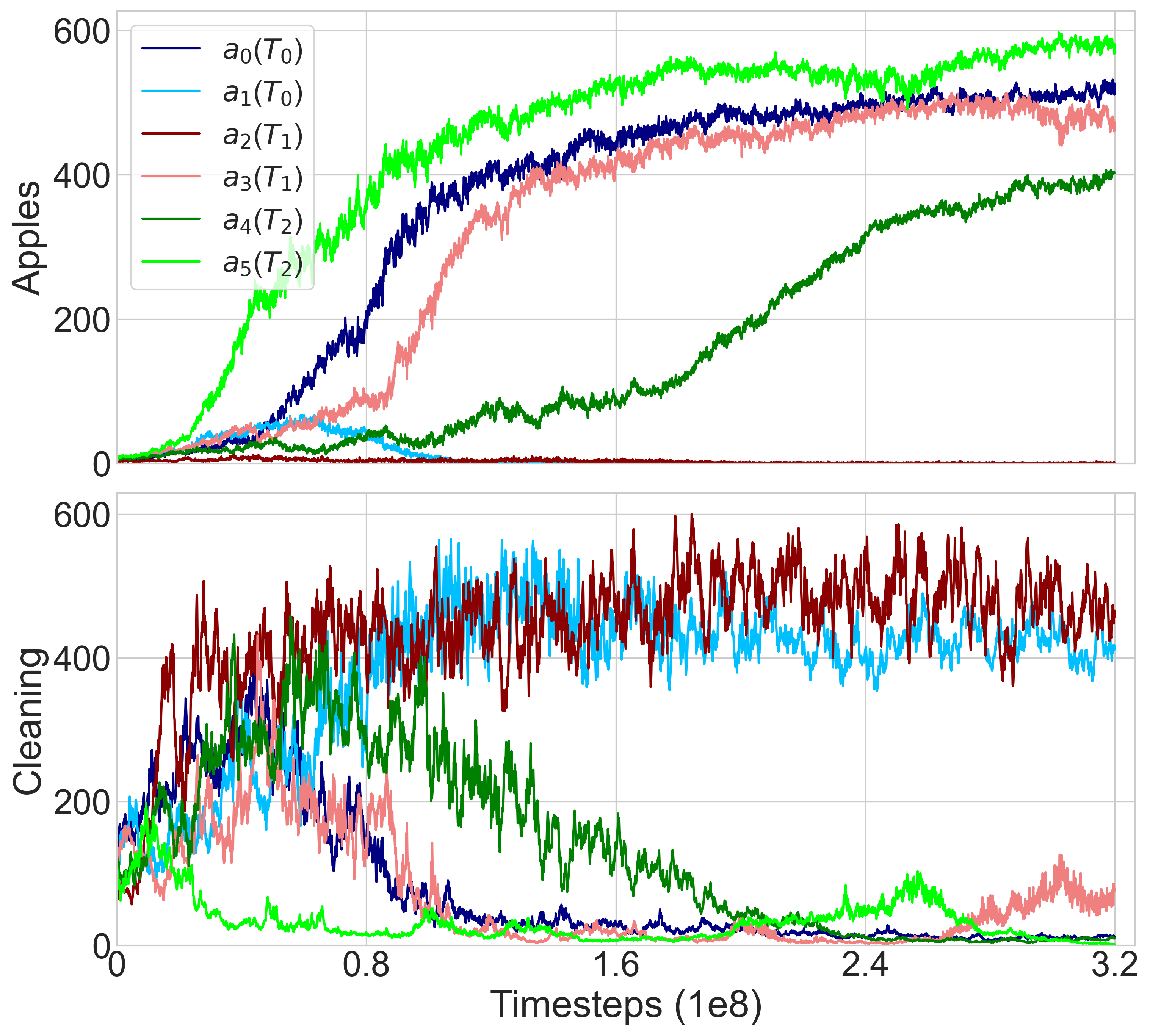}
    % \vspace{-20pt}
    \caption{Amount of apples consumed (top) and cleaning beam actions (bottom) by each agent for one trial of the credo-tuning experiment with agents initialized with system-focused credo (green line in Figures~\ref{fig:reward_curve} and~\ref{fig:gini_curve}). Agents are labeled so that $a_0 (T_0)$ is agent 0 on team 0. Teammates are colored with different shades of the same color. Whereas system-focused agents converge to a joint policy of three apple pickers and three cleaning agents, credo-tuning agents recover the better joint policy of four apple pickers and two cleaning agents autonomously (same as fully team-focused agents) and generate more reward (Figure~\ref{fig:reward_curve}).}
    \label{fig:metrics_curves}
\end{figure}

\begin{figure}[t]
    \centering
    \includegraphics[width=\linewidth]{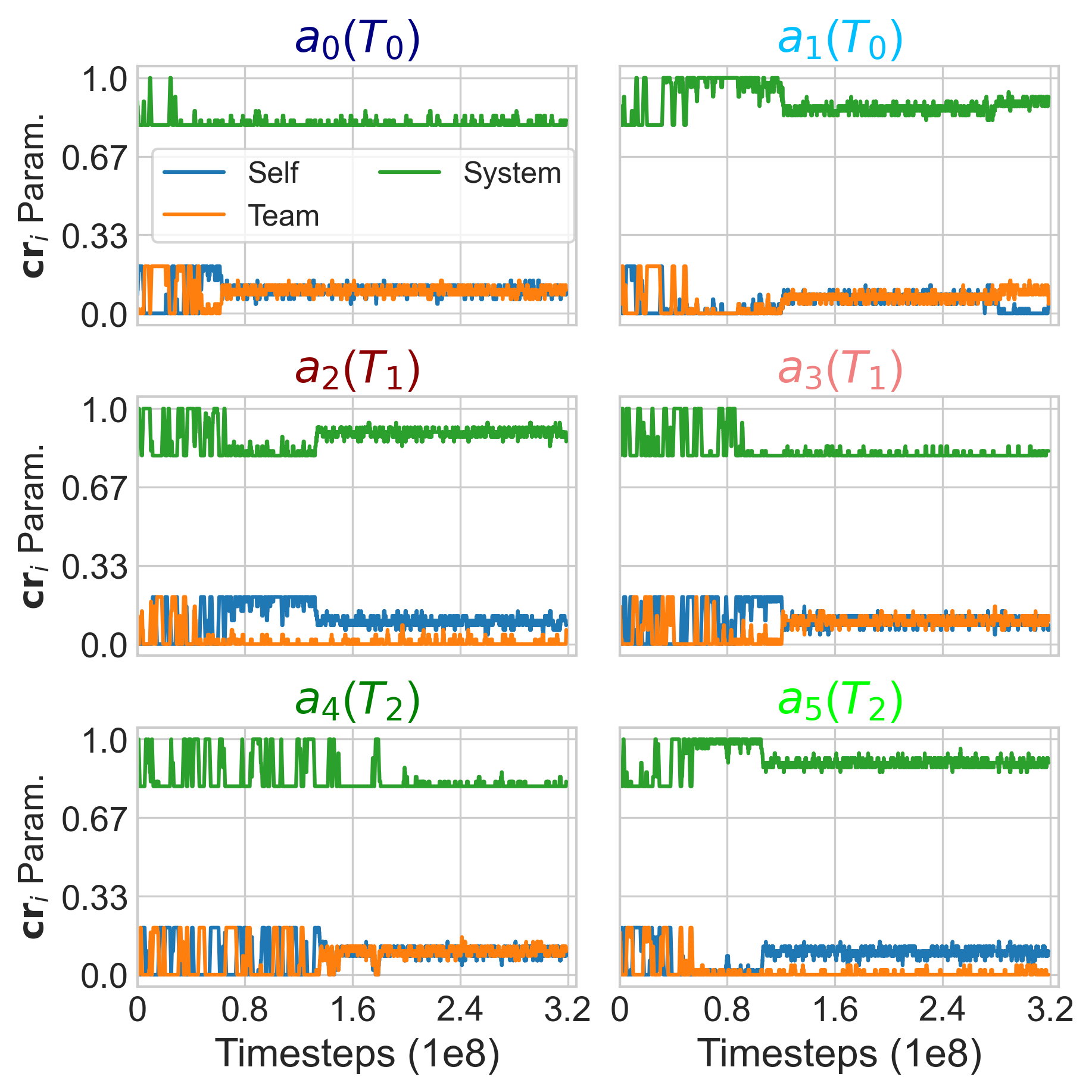}
    % \vspace{-20pt}
    \caption{Credos of all six agents over time in the same credo-tuning trial as Figure~\ref{fig:metrics_curves}. Each plot shows the credo parameters for a different agent shown in Figure~\ref{fig:metrics_curves}. Each $y$-axis represents credo parameter space and each $x$-axis represents timesteps. We observe heterogeneous credo parameters emerge across the population; however, $a_4$ becomes more self- and team-focused as it switches roles to become an apple picking agent.}
    % Their teammate being a river cleaning agent (no environment reward) likely causing the change in behavior.}
    \label{fig:credos}
\end{figure}

Since certain roles in the environment do not produce reward and teammates are able to define different credos, it is important to consider population equality to examine if tuning credo leads to significant inequality among the population.
% , or how evenly reward is distributed among a population of agents.
We model population reward equality as the inverse Gini index, similar to past work~\cite{McKee2020SocialDA,Radke2022Exploring}:

\begin{equation}
    Equality = 1 - \frac{\sum_{i=0}^{N} \sum_{j=0}^{N} |R_{i}^{\mathbf{cr}} - R_{j}^{\mathbf{cr}}|}{2N^{2} \overline{R^{\mathbf{cr}}}},
\end{equation}

\noindent
where values closer to 1 represent more equality.
Figure~\ref{fig:gini_curve} shows our equality results, where the $y$-axis shows the mean inverse Gini index with 95\% confidence intervals and the $x$-axis is the number of timesteps.
Since the static system-focus scenario defines agents to fully share rewards, the inverse Gini index is always equal to 1.
After some initial learning, we find that the credo-tuning agents converge to a setting where the population has higher mean equality than the static team-focused setting.
While this is likely impacted by the credo intialization and is worthy of further exploration, we find that credo-tuning agents discover a setting that achieves high reward while maintaining high equality across the population.
% Note that these results come from only a single experiment and more trials are needed to fully understand the variance of our results.

\subsubsection{Division of Labor}
\label{subsec:div_of_labor}

We now analyze the credo-tuning experiment specifically.
Figure~\ref{fig:metrics_curves} shows the amount of apples consumed (top) and cleaning beam actions (bottom) by each credo-tuning agent in one trial where the agents are initialized to be fully system-focused (green line in Figures~\ref{fig:reward_curve} and~\ref{fig:gini_curve}).
Despite being initialized as system-focused, these agents have team membership to one of three teams ($T_0$, $T_1$, or $T_2$) to modify their credo towards.
Agents are labeled so that $a_0 (T_0)$ represents agent 0 on team 0 and teammates in the plots are colored with different shades of the same color.

Similar to the known result of static system-focused agents shown in past work~\cite{radke2022importance}, the agents in the credo-tuning experiment initially specialize into roles of three apple picker and three river cleaning agents.
However, the advantage of agents being able to tune their credo causes the $a_4 (T_2)$ agent to learn to pick apples in the second half of the experiment.
This recovers the global joint policy of four apple picker agents and two river cleaning agents (joint policy of the static team-focused agents) despite agents being initialized with full system-focused credo.
This causes an increase in mean population reward from the static system-focused scenario towards the full team-focused scenario.
While the mean population reward level of team-focused agents is not quite reached, these agents recover the same global joint policy; thus, while we are unable to make certain claims, perhaps longer training time would see convergence to the reward level of the team-focused population (blue in Figure~\ref{fig:reward_curve}) given this joint policy.
% If this behavior is sustained and the experiment were ran for longer, we expect the mean population reward in Figure~\ref{fig:reward_curve} for credo-tuning agents (green) to converge to the static team-focused scenario (blue).

\subsubsection{Tuned Credo Parameters}
\label{subsec:tuned_credo}

% \todo{Mention updated libraries make curves not comparable to past results, but overall results are the same.}

% Figure~\ref{fig:metrics_curves} shows how the high-level credo policy modifies each agent's credo parameters over time.
% and compare this with the behavior learned.
Figure~\ref{fig:credos} shows how the credo parameters for each agent in the trial shown in Figure~\ref{fig:metrics_curves}, modified by each agents' high-level credo policy.
Each plot is titled and colored according to the agent's label and color in Figure~\ref{fig:metrics_curves}.
The $y$-axis of each plot shows the credo parameter values and the $x$-axis of each plot shows timesteps of the experiment.
The results for each credo parameter are a sliding window mean of every 10 samples; thus, some results appear between two discrete credo steps (such as 0.1 being between 0.0 and 0.2).

Figure~\ref{fig:credos} shows that two teammates that converge to complimentary roles of one river cleaner and one apple picker, $a_0$ and $a_1$ (blue; $T_0$), maintain periods of non-zero team focus.
This allows the agents to share some of the reward gained by their teammate while sharing the majority of their apples through the system-focus reward channel.
The other team that divides labor between two roles over the entire experiment, $a_2$ and $a_3$ (red; $T_1$), have heterogeneous credo parameters amongst their team.
While the cleaning agent $a_2$ maintains higher system-focus, the apple picking agent $a_3$ has slightly higher self-focus to keep some amount of the reward they collect to themselves.
The agent that changes roles to become an apple picker, $a_4$ on $T_2$, maintains a period of being self- and team-focused, before $1.5 \times 10^8$ timesteps.
At this time, their teammate ($a_5 (T_2)$) develops a credo where they do not share rewards through the team parameter, instead maintaining high system-focus before becoming slightly self-focused.
After a period where their teammate is not contributing to the team reward when $a_4$ is slightly team-focused, the agent switches behaviors to become an apple picking agent.
This may indicate why $a_4$ becomes an apple picker with some amount of self-focus (i.e., increasing their personal reward).

These results show how our framework allows for diverse group alignments to be learned.
In turn, these learned heterogeneous alignments lead to agents recovering a globally better joint policy while maintaining high equality.

% also learns to pick apples and tunes its credo to become slightly team-focused.
% We find that agents on team 1 ($T_1$, red) maintain high degree of system-focus, although $a_2$ goes through a period of being slightly team-focused before returning to be fully system-focus.

% Team 2 ($T_2$, green) is the only team where agents learn the same pattern of credo parameters and this team also experienced a behavior shift mid-experiment.
% Just before $1.0 \times 10^8$ timesteps, $a_4$ maintains a period of slight team-focus; however, its teammate ($a_5$) is also cleaning the river and not collecting any environmental reward.
% At this point, $a_4$ switches behavior and becomes an apple picking agent (shown in Figure~\ref{fig:metrics_curves}).
% Both agents in $T_2$ then become slightly team-focused and share a fraction of the reward gained by $a_4$ between themselves.
% These results show how our framework allows for diverse group alignments to be learned.
% In turn, these learned heterogeneous alignments lead to agents recovering a globally better joint policy while maintaining high equality.

\section{Future Work}
\label{sec:future_work}

% Talk about nmmo and deep credo agent.

This paper presented an initial framework and evaluation of credo-tuning agents as a preliminary proof of concept.
We will expand this work in the two main areas of experimental evaluation and model design.

\subsection{Experimental Future Work}
\label{subsec:experiments_future}
We plan on performing a more extensive empirical evaluation of our framework.
We plan on designing more specific experimental scenarios such as initializing agents to be fully team-focused, self-focused, or randomly distributed across different credo parameters.These experiments will provide insight into the convergence properties of credo-tuning agents (i.e., if there is one or multiple convergence points in credo parameters for different team structures).
Furthermore, allowing agents to belong to multiple teams, and self-regulate their credo for each specific team, is an area of future research that could provide insights into dynamic team membership or multi-group alignment~\cite{tilbury2022identity}. We also plan on expanding our evaluation to the Neural MMO (NMMO)~\cite{suarez2019neural} environment.
NMMO is a large, customizable, and partially observable multiagent environment that supports foraging and exploration.
% Similar to Cleanup, agents are partially observable and have a variety of actions.
Implementing credo-tuning agents in NMMO will help understand the connections between how agents modify their credo in hunter-gatherer-type societies and in scenarios where the most beneficial joint policy may be unknown.

\subsection{Model Design}
\label{sec:model_future}
We plan to improve and expand our model design.
We implemented the credo policy using $Q$-Learning with discrete credo step sizes of 0.2 to reduce sample and state/action space complexity for the high-level policy.
The state space consisted of a finite set of discrete states and had a finite set of discrete actions.
However, recent work in single-agent RL has shown significant advances in continuous control problems where both the state and action spaces consist of high-dimensional continuous values~\cite{wurman2022outracing}.
Other work has shown the ability to decompose the Bellman equation into a vectorized representation to learn the value of different reward parts~\cite{macglashan2022value}.
We plan on expanding the development of the high-level credo policy to incorporate continuous control designs to learn the value of various reward components (self, team, or system).
Furthermore, we plan to expand the action space to directly output credo parameters instead of discrete modifications.
Progress in this direction will expand reward function decomposition to the multiagent scenario while allowing for significantly more credo parameter combinations among the population.

\section{Conclusion}

This work presented the design of credo-tuning agent framework influenced by hierarchical reinforcement learning (HRL) and meta-learning.
The dual-tiered architecture draws inspiration from the multi-layer optimization problems of HRL; however, the influences on learning dynamics at each layer are more similar to a meta-learning problem.
In meta-learning, \emph{learning to learn} is the idea in which an agent learns at two levels, each associated with different time scales~\cite{thrun1998lifelong,santoro2016meta}.
% The low-level behavioral policy learns how to behave in the environment, converge to roles within groups it belongs to, and how to interact with other agents.
The ability for the behavioral policy to learn effective policies and roles among a group is guided by their reward function's feedback signals, shaped by their credo.
% the environment's team structure and agent's credo parameters for various groups.
The high-level credo policy updates these credo parameters at a slower timescale, changing the low-level behavioral policies optimization landscape and guiding the learning process through different reward components.
% for the low-level behavioral policy by modifying the agent's credo parameters.
% Since an agent's credo parameters are updated by their high-level credo-tuning policy operating at a slower timescale, the high-level policy essentially shapes the learning problem for the low-level policy over time.
% that operates at a slower time scale across episodes.
% The high-level policy shapes the optimization landscape for the low-level behavioral policy, while the low-level policies behavior shapes the reward signal for the high-level policy.
% Thus, a successful learning agent's high-level policy would learn favorable credo parameters that allow the low-level behavioral policy to learn -- learning to learn.
Dynamically tuning credo allows this meta-learning problem to evolve online while maintaining the decentralized aspects of individual learning agents.
% In environments with sub-optimal team structures, such as teams that are too small or too large, individual learning agents could converge to joint distributions of credo parameters that guide the population towards favorable joint policies.

The goal of this work is to allow decentralized agents to self-regulate their credo to overcome sub-optimal initializations of credo or team structures and recover favorable policies.
While previous work has shown how team structure has a significant impact on the policies that agents learn, discovering the structure that guides agents towards globally favorable results may be a hard domain dependent problem.
% This work is a first step towards designing agents with the ability to recover favorable policies despite sub-optimal team structures.
Our preliminary results have shown how our multi-tiered learning architecture can allow agents to achieve more globally favorable results despite being initialized in a known sub-optimal configuration.
The broader implications of this work allow agents to autonomously recover the learning benefits of teams and group alignment in any environment.
This mitigates the burden of researchers or practitioners having to engineer team structures or credo in settings where favorable configurations may be unknown.

%%%%%%%%%%%%%%%%%%%%%%%%%%%%%%%%%%%%%%%%%%%%%%%%%%%%%%%%%%%%%%%%%%%%%%%%

%%% The next two lines define, first, the bibliography style to be 
%%% applied, and, second, the bibliography file to be used.

\bibliographystyle{ACM-Reference-Format} 
% \bibliography{references}


\begin{thebibliography}{26}

%%% the Web address.

\ifx \showCODEN    \undefined \def \showCODEN     #1{\unskip}     \fi
\ifx \showDOI      \undefined \def \showDOI       #1{#1}\fi
\ifx \showISBNx    \undefined \def \showISBNx     #1{\unskip}     \fi
\ifx \showISBNxiii \undefined \def \showISBNxiii  #1{\unskip}     \fi
\ifx \showISSN     \undefined \def \showISSN      #1{\unskip}     \fi
\ifx \showLCCN     \undefined \def \showLCCN      #1{\unskip}     \fi
\ifx \shownote     \undefined \def \shownote      #1{#1}          \fi
\ifx \showarticletitle \undefined \def \showarticletitle #1{#1}   \fi
\ifx \showURL      \undefined \def \showURL       {\relax}        \fi
% invisible to TeX
\providecommand\bibfield[2]{#2}
\providecommand\bibinfo[2]{#2}
\providecommand\natexlab[1]{#1}
\providecommand\showeprint[2][]{arXiv:#2}

\bibitem[\protect\citeauthoryear{Carter, Asencio, Trainer, DeChurch, Kanfer,
  and Zaccaro}{Carter et~al\mbox{.}}{2019}]%
        {Carter2019BestPF}
\bibfield{author}{\bibinfo{person}{D.~R. Carter}, \bibinfo{person}{R. Asencio},
  \bibinfo{person}{Hayley~M. Trainer}, \bibinfo{person}{Leslie~A. DeChurch},
  \bibinfo{person}{R. Kanfer}, {and} \bibinfo{person}{S. Zaccaro}.}
  \bibinfo{year}{2019}\natexlab{}.
\newblock \showarticletitle{Best Practices for Researchers Working in Multiteam
  Systems}.
\newblock


\bibitem[\protect\citeauthoryear{Dafoe, Bachrach, Hadfield, Horvitz, Larson,
  and Graepel}{Dafoe et~al\mbox{.}}{2021}]%
        {DafoeNature2021}
\bibfield{author}{\bibinfo{person}{Allan Dafoe}, \bibinfo{person}{Yoram
  Bachrach}, \bibinfo{person}{Gillian Hadfield}, \bibinfo{person}{Eric
  Horvitz}, \bibinfo{person}{Kate Larson}, {and} \bibinfo{person}{Thore
  Graepel}.} \bibinfo{year}{2021}\natexlab{}.
\newblock \showarticletitle{Cooperative {AI}: machines must learn to find
  common ground}.
\newblock \bibinfo{journal}{\emph{Nature}}  \bibinfo{volume}{593}
  (\bibinfo{year}{2021}), \bibinfo{pages}{33--36}.
\newblock


\bibitem[\protect\citeauthoryear{Dafoe, Hughes, Bachrach, Collins, McKee,
  Leibo, Larson, and Graepel}{Dafoe et~al\mbox{.}}{2020}]%
        {Dafoe2020OpenPI}
\bibfield{author}{\bibinfo{person}{A. Dafoe}, \bibinfo{person}{Edward Hughes},
  \bibinfo{person}{Yoram Bachrach}, \bibinfo{person}{Tantum Collins},
  \bibinfo{person}{Kevin~R. McKee}, \bibinfo{person}{Joel~Z. Leibo},
  \bibinfo{person}{K. Larson}, {and} \bibinfo{person}{T. Graepel}.}
  \bibinfo{year}{2020}\natexlab{}.
\newblock \showarticletitle{Open Problems in Cooperative AI}.
\newblock \bibinfo{journal}{\emph{ArXiv}}  \bibinfo{volume}{abs/2012.08630}
  (\bibinfo{year}{2020}).
\newblock


\bibitem[\protect\citeauthoryear{Dunbar}{Dunbar}{1993}]%
        {dunbar1993coevolution}
\bibfield{author}{\bibinfo{person}{Robin~IM Dunbar}.}
  \bibinfo{year}{1993}\natexlab{}.
\newblock \showarticletitle{Coevolution of neocortical size, group size and
  language in humans}.
\newblock \bibinfo{journal}{\emph{Behavioral and brain sciences}}
  \bibinfo{volume}{16}, \bibinfo{number}{4} (\bibinfo{year}{1993}),
  \bibinfo{pages}{681--694}.
\newblock


\bibitem[\protect\citeauthoryear{Durugkar, Liebman, and Stone}{Durugkar
  et~al\mbox{.}}{2020}]%
        {Durugkar2020BalancingIP}
\bibfield{author}{\bibinfo{person}{Ishan Durugkar}, \bibinfo{person}{E.
  Liebman}, {and} \bibinfo{person}{P. Stone}.} \bibinfo{year}{2020}\natexlab{}.
\newblock \showarticletitle{Balancing Individual Preferences and Shared
  Objectives in Multiagent Reinforcement Learning}. In
  \bibinfo{booktitle}{\emph{IJCAI}}.
\newblock


\bibitem[\protect\citeauthoryear{Gemp, McKee, Everett,
  Du{\'e}{\~n}ez-Guzm{\'a}n, Bachrach, Balduzzi, and Tacchetti}{Gemp
  et~al\mbox{.}}{2022}]%
        {gemp2020d3c}
\bibfield{author}{\bibinfo{person}{Ian Gemp}, \bibinfo{person}{Kevin~R McKee},
  \bibinfo{person}{Richard Everett}, \bibinfo{person}{Edgar~A
  Du{\'e}{\~n}ez-Guzm{\'a}n}, \bibinfo{person}{Yoram Bachrach},
  \bibinfo{person}{David Balduzzi}, {and} \bibinfo{person}{Andrea Tacchetti}.}
  \bibinfo{year}{2022}\natexlab{}.
\newblock \showarticletitle{D3C: Reducing the Price of Anarchy in Multi-Agent
  Learning}.
\newblock \bibinfo{journal}{\emph{Proceedings of the 21st International
  Conference on Autonomous Agents and MultiAgent Systems}}
  (\bibinfo{year}{2022}).
\newblock


\bibitem[\protect\citeauthoryear{Grosz and Kraus}{Grosz and Kraus}{1996}]%
        {Grosz1996CollaborativePF}
\bibfield{author}{\bibinfo{person}{B. Grosz} {and} \bibinfo{person}{S. Kraus}.}
  \bibinfo{year}{1996}\natexlab{}.
\newblock \showarticletitle{Collaborative Plans for Complex Group Action}.
\newblock \bibinfo{journal}{\emph{Artif. Intell.}}  \bibinfo{volume}{86}
  (\bibinfo{year}{1996}), \bibinfo{pages}{269--357}.
\newblock


\bibitem[\protect\citeauthoryear{Leibo, Zambaldi, Lanctot, Marecki, and
  Graepel}{Leibo et~al\mbox{.}}{2017}]%
        {Leibo2017MultiagentRL}
\bibfield{author}{\bibinfo{person}{J.~Z. Leibo}, \bibinfo{person}{V. Zambaldi},
  \bibinfo{person}{M. Lanctot}, \bibinfo{person}{J. Marecki}, {and}
  \bibinfo{person}{T. Graepel}.} \bibinfo{year}{2017}\natexlab{}.
\newblock \showarticletitle{Multi-agent Reinforcement Learning in Sequential
  Social Dilemmas}. In \bibinfo{booktitle}{\emph{AAMAS}}.
\newblock


\bibitem[\protect\citeauthoryear{MacGlashan, Archer, Devlic, Seno, Sherstan,
  Wurman, and Stone}{MacGlashan et~al\mbox{.}}{2022}]%
        {macglashan2022value}
\bibfield{author}{\bibinfo{person}{James MacGlashan}, \bibinfo{person}{Evan
  Archer}, \bibinfo{person}{Alisa Devlic}, \bibinfo{person}{Takuma Seno},
  \bibinfo{person}{Craig Sherstan}, \bibinfo{person}{Peter~R Wurman}, {and}
  \bibinfo{person}{Peter Stone}.} \bibinfo{year}{2022}\natexlab{}.
\newblock \showarticletitle{Value Function Decomposition for Iterative Design
  of Reinforcement Learning Agents}.
\newblock \bibinfo{journal}{\emph{NeurIPS}} (\bibinfo{year}{2022}).
\newblock


\bibitem[\protect\citeauthoryear{Macke, Mirsky, and Stone}{Macke
  et~al\mbox{.}}{2021}]%
        {Macke2021ExpectedVO}
\bibfield{author}{\bibinfo{person}{W. Macke}, \bibinfo{person}{R. Mirsky},
  {and} \bibinfo{person}{P. Stone}.} \bibinfo{year}{2021}\natexlab{}.
\newblock \showarticletitle{Expected Value of Communication for Planning in Ad
  Hoc Teamwork}. In \bibinfo{booktitle}{\emph{AAAI-21}}.
\newblock


\bibitem[\protect\citeauthoryear{McKee, Gemp, McWilliams,
  Du{\'e}{\~n}ez-Guzm{\'a}n, Hughes, and Leibo}{McKee et~al\mbox{.}}{2020}]%
        {McKee2020SocialDA}
\bibfield{author}{\bibinfo{person}{Kevin~R. McKee}, \bibinfo{person}{I. Gemp},
  \bibinfo{person}{Brian McWilliams}, \bibinfo{person}{Edgar~A.
  Du{\'e}{\~n}ez-Guzm{\'a}n}, \bibinfo{person}{Edward Hughes}, {and}
  \bibinfo{person}{Joel~Z. Leibo}.} \bibinfo{year}{2020}\natexlab{}.
\newblock \showarticletitle{Social Diversity and Social Preferences in
  Mixed-Motive Reinforcement Learning}.
\newblock \bibinfo{journal}{\emph{AAMAS}} (\bibinfo{year}{2020}).
\newblock


\bibitem[\protect\citeauthoryear{Porck, Matta, Hollenbeck, Oh, Lanaj, and
  Lee}{Porck et~al\mbox{.}}{2019}]%
        {Porck2019SocialII}
\bibfield{author}{\bibinfo{person}{J. Porck}, \bibinfo{person}{Fadel~K Matta},
  \bibinfo{person}{J. Hollenbeck}, \bibinfo{person}{Jo~K. Oh},
  \bibinfo{person}{Klodiana Lanaj}, {and} \bibinfo{person}{S. Lee}.}
  \bibinfo{year}{2019}\natexlab{}.
\newblock \showarticletitle{Social Identification in Multiteam Systems: The
  Role of Depletion and Task Complexity}.
\newblock \bibinfo{journal}{\emph{Academy of Management Journal}}
  \bibinfo{volume}{62} (\bibinfo{year}{2019}), \bibinfo{pages}{1137--1162}.
\newblock


\bibitem[\protect\citeauthoryear{Radke, Larson, and Brecht}{Radke
  et~al\mbox{.}}{2022}]%
        {Radke2022Exploring}
\bibfield{author}{\bibinfo{person}{David Radke}, \bibinfo{person}{Kate Larson},
  {and} \bibinfo{person}{Tim Brecht}.} \bibinfo{year}{2022}\natexlab{}.
\newblock \showarticletitle{Exploring the Benefits of Teams in Multiagent
  Learning}. In \bibinfo{booktitle}{\emph{IJCAI}}.
\newblock


\bibitem[\protect\citeauthoryear{Radke, Larson, and Brecht}{Radke
  et~al\mbox{.}}{2023}]%
        {radke2022importance}
\bibfield{author}{\bibinfo{person}{David Radke}, \bibinfo{person}{Kate Larson},
  {and} \bibinfo{person}{Tim Brecht}.} \bibinfo{year}{2023}\natexlab{}.
\newblock \showarticletitle{The Importance of Credo in Multiagent Learning}.
\newblock \bibinfo{journal}{\emph{Proceedings of the 22nd International
  Conference on Autonomous Agents and MultiAgent Systems}}
  (\bibinfo{year}{2023}).
\newblock


\bibitem[\protect\citeauthoryear{Santoro, Bartunov, Botvinick, Wierstra, and
  Lillicrap}{Santoro et~al\mbox{.}}{2016}]%
        {santoro2016meta}
\bibfield{author}{\bibinfo{person}{Adam Santoro}, \bibinfo{person}{Sergey
  Bartunov}, \bibinfo{person}{Matthew Botvinick}, \bibinfo{person}{Daan
  Wierstra}, {and} \bibinfo{person}{Timothy Lillicrap}.}
  \bibinfo{year}{2016}\natexlab{}.
\newblock \showarticletitle{Meta-learning with memory-augmented neural
  networks}. In \bibinfo{booktitle}{\emph{International conference on machine
  learning}}. PMLR, \bibinfo{pages}{1842--1850}.
\newblock


\bibitem[\protect\citeauthoryear{Schulman, Wolski, Dhariwal, Radford, and
  Klimov}{Schulman et~al\mbox{.}}{2017}]%
        {PPO2017}
\bibfield{author}{\bibinfo{person}{John Schulman}, \bibinfo{person}{Filip
  Wolski}, \bibinfo{person}{Prafulla Dhariwal}, \bibinfo{person}{Alec Radford},
  {and} \bibinfo{person}{Oleg Klimov}.} \bibinfo{year}{2017}\natexlab{}.
\newblock \showarticletitle{Proximal Policy Optimization Algorithms}.
\newblock \bibinfo{journal}{\emph{CoRR}} (\bibinfo{year}{2017}).
\newblock


\bibitem[\protect\citeauthoryear{Stone, Kaminka, Kraus, and Rosenschein}{Stone
  et~al\mbox{.}}{2010}]%
        {Stone2010AdHA}
\bibfield{author}{\bibinfo{person}{P. Stone}, \bibinfo{person}{G. Kaminka},
  \bibinfo{person}{S. Kraus}, {and} \bibinfo{person}{J. Rosenschein}.}
  \bibinfo{year}{2010}\natexlab{}.
\newblock \showarticletitle{Ad Hoc Autonomous Agent Teams: Collaboration
  without Pre-Coordination}. In \bibinfo{booktitle}{\emph{AAAI}}.
\newblock


\bibitem[\protect\citeauthoryear{Suarez, Du, Isola, and Mordatch}{Suarez
  et~al\mbox{.}}{2019}]%
        {suarez2019neural}
\bibfield{author}{\bibinfo{person}{Joseph Suarez}, \bibinfo{person}{Yilun Du},
  \bibinfo{person}{Phillip Isola}, {and} \bibinfo{person}{Igor Mordatch}.}
  \bibinfo{year}{2019}\natexlab{}.
\newblock \showarticletitle{Neural MMO: A massively multiagent game environment
  for training and evaluating intelligent agents}.
\newblock \bibinfo{journal}{\emph{arXiv preprint arXiv:1903.00784}}
  (\bibinfo{year}{2019}).
\newblock


\bibitem[\protect\citeauthoryear{Sundstrom, Meuse, and Futrell}{Sundstrom
  et~al\mbox{.}}{1990}]%
        {Sundstrom1990WorkTA}
\bibfield{author}{\bibinfo{person}{E. Sundstrom}, \bibinfo{person}{K.~P.~D.
  Meuse}, {and} \bibinfo{person}{D. Futrell}.} \bibinfo{year}{1990}\natexlab{}.
\newblock \showarticletitle{Work teams: Applications and effectiveness.}
\newblock \bibinfo{journal}{\emph{American Psychologist}}  \bibinfo{volume}{45}
  (\bibinfo{year}{1990}), \bibinfo{pages}{120--133}.
\newblock


\bibitem[\protect\citeauthoryear{Tambe}{Tambe}{1997}]%
        {Tambe1997TowardsFT}
\bibfield{author}{\bibinfo{person}{Milind Tambe}.}
  \bibinfo{year}{1997}\natexlab{}.
\newblock \showarticletitle{Towards Flexible Teamwork}.
\newblock \bibinfo{journal}{\emph{Journal of Artificial Intelligence Research}}
   \bibinfo{volume}{7} (\bibinfo{year}{1997}), \bibinfo{pages}{83--124}.
\newblock


\bibitem[\protect\citeauthoryear{Thrun}{Thrun}{1998}]%
        {thrun1998lifelong}
\bibfield{author}{\bibinfo{person}{Sebastian Thrun}.}
  \bibinfo{year}{1998}\natexlab{}.
\newblock \showarticletitle{Lifelong learning algorithms}.
\newblock In \bibinfo{booktitle}{\emph{Learning to learn}}.
  \bibinfo{publisher}{Springer}, \bibinfo{pages}{181--209}.
\newblock


\bibitem[\protect\citeauthoryear{Tilbury and Hoey}{Tilbury and Hoey}{2022}]%
        {tilbury2022identity}
\bibfield{author}{\bibinfo{person}{Kyle Tilbury} {and} \bibinfo{person}{Jesse
  Hoey}.} \bibinfo{year}{2022}\natexlab{}.
\newblock \showarticletitle{Identity and Dynamic Teams in Social Dilemmas}.
\newblock \bibinfo{journal}{\emph{arXiv preprint arXiv:2208.03293}}
  (\bibinfo{year}{2022}).
\newblock


\bibitem[\protect\citeauthoryear{Vinitsky, Jaques, Leibo, Castenada, and
  Hughes}{Vinitsky et~al\mbox{.}}{2019}]%
        {SSDOpenSource}
\bibfield{author}{\bibinfo{person}{Eugene Vinitsky}, \bibinfo{person}{Natasha
  Jaques}, \bibinfo{person}{Joel Leibo}, \bibinfo{person}{Antonio Castenada},
  {and} \bibinfo{person}{Edward Hughes}.} \bibinfo{year}{2019}\natexlab{}.
\newblock \bibinfo{title}{An Open Source Implementation of Sequential Social
  Dilemma Games}.
\newblock
  \bibinfo{howpublished}{\url{https://github.com/eugenevinitsky/sequential_social_dilemma_games/issues/182}}.
\newblock
\newblock
\shownote{GitHub repository.}


\bibitem[\protect\citeauthoryear{Watkins and Dayan}{Watkins and Dayan}{1992}]%
        {watkins1992q}
\bibfield{author}{\bibinfo{person}{Christopher~JCH Watkins} {and}
  \bibinfo{person}{Peter Dayan}.} \bibinfo{year}{1992}\natexlab{}.
\newblock \showarticletitle{Q-learning}.
\newblock \bibinfo{journal}{\emph{Machine learning}}  \bibinfo{volume}{8}
  (\bibinfo{year}{1992}), \bibinfo{pages}{279--292}.
\newblock


\bibitem[\protect\citeauthoryear{Wijnmaalen, Voordijk, Rietjens, and
  Dewulf}{Wijnmaalen et~al\mbox{.}}{2019}]%
        {Wijnmaalen2019IntergroupBI}
\bibfield{author}{\bibinfo{person}{Julia Wijnmaalen}, \bibinfo{person}{H.
  Voordijk}, \bibinfo{person}{S. Rietjens}, {and} \bibinfo{person}{G. Dewulf}.}
  \bibinfo{year}{2019}\natexlab{}.
\newblock \showarticletitle{Intergroup behavior in military multiteam systems}.
\newblock \bibinfo{journal}{\emph{Human Relations}}  \bibinfo{volume}{72}
  (\bibinfo{year}{2019}), \bibinfo{pages}{1081 -- 1104}.
\newblock


\bibitem[\protect\citeauthoryear{Wurman, Barrett, Kawamoto, MacGlashan,
  Subramanian, Walsh, Capobianco, Devlic, Eckert, Fuchs, et~al\mbox{.}}{Wurman
  et~al\mbox{.}}{2022}]%
        {wurman2022outracing}
\bibfield{author}{\bibinfo{person}{Peter~R Wurman}, \bibinfo{person}{Samuel
  Barrett}, \bibinfo{person}{Kenta Kawamoto}, \bibinfo{person}{James
  MacGlashan}, \bibinfo{person}{Kaushik Subramanian}, \bibinfo{person}{Thomas~J
  Walsh}, \bibinfo{person}{Roberto Capobianco}, \bibinfo{person}{Alisa Devlic},
  \bibinfo{person}{Franziska Eckert}, \bibinfo{person}{Florian Fuchs},
  {et~al\mbox{.}}} \bibinfo{year}{2022}\natexlab{}.
\newblock \showarticletitle{Outracing champion Gran Turismo drivers with deep
  reinforcement learning}.
\newblock \bibinfo{journal}{\emph{Nature}} \bibinfo{volume}{602},
  \bibinfo{number}{7896} (\bibinfo{year}{2022}), \bibinfo{pages}{223--228}.
\newblock


\end{thebibliography}

%%%%%%%%%%%%%%%%%%%%%%%%%%%%%%%%%%%%%%%%%%%%%%%%%%%%%%%%%%%%%%%%%%%%%%%%

\end{document}